\documentclass[conference]{IEEEtran}
\usepackage{graphicx}

\begin{document}

\title{Conjugate Gradient Acceleration of\\ Non-Linear Smoothing Filters
}

\author{\IEEEauthorblockN{Andrew Knyazev}
\IEEEauthorblockA{Mitsubishi Electric Research Laboratories (MERL)\\
201 Broadway, 8th floor, Cambridge, MA 02139, USA\\
Email: knyazev@merl.com}
\and
\IEEEauthorblockN{Alexander Malyshev}
\IEEEauthorblockA{Mitsubishi Electric Research Laboratories (MERL)\\
201 Broadway, 8th floor, Cambridge, MA 02139, USA\\
Email: malyshev@merl.com}
}



\maketitle

\begin{abstract}
The most efficient signal edge-preserving smoothing filters, e.g., for denoising, are non-linear.
Thus, their acceleration is challenging and is often performed in practice by tuning filter parameters,
such as by increasing the width of the local smoothing neighborhood, resulting in more aggressive
smoothing of a single sweep at the cost of increased edge blurring.
We propose an alternative technology, accelerating the original filters without tuning,
by running them through a special conjugate gradient method, not affecting their quality.
The filter non-linearity is dealt with by careful freezing and restarting.
Our initial numerical experiments on toy one-dimensional signals demonstrate
20x acceleration of the classical bilateral filter and 3-5x acceleration
of the recently developed guided filter. 
\end{abstract}

\begin{keywords}
conjugate gradient algorithm, edge-preserving denoising, low-pass filters
\end{keywords}

%
\IEEEpeerreviewmaketitle

\section{Introduction}
\label{sec:intro}

This paper\footnote{Accepted to the 3rd IEEE GlobalSIP Conference 2015}
is concerned with noise removal from a given noisy signal, which is a basic problem
in signal processing, with many applications, e.g.,\ in image denoising \cite{M13}.
Modern denoising algorithms preserve signal details while removing most of the noise.
A very popular denoising filter is the bilateral filter (BF), which
smooths signals while preserving edges, by taking the weighted average of the nearby pixels.
The weights depend on both the spatial distance between the sampling locations and
similarity between signal values, thus providing local adaptivity to the input signal.
Bilateral filtering has initially been proposed in \cite{TM98} as an intuitive tool
without theoretical justification. Since then, connections between BF and other well-known
filtering techniques such as anisotropic diffusion, weighted least squares, Bayesian methods,
kernel regression and non-local means have been explored; see, e.g., survey \cite{PKTD09}.

We make use of the graph-based framework for signal analysis developed in \cite{TKMV14,GNO13},
where polynomial low-pass filters based on the BF coefficients are proposed.
A nice introduction to signal processing on graphs is found in \cite{FOSNV13}.

A single application of BF can be interpreted as a vertex domain transform on a graph
with pixels as vertices, intensity values of each node as the graph signal, and filter coefficients
as link weights that capture the similarity between nodes. The BF transform is a special nonlinear
anisotropic diffusion, cf. \cite{PM87,PM90}, determined by the entries of the graph Laplacian matrix,
which are related to the BF weights. The eigenvectors and eigenvalues of the graph Laplacian matrix
allow us to extend the Fourier analysis to the graph signals or images as in \cite{FOSNV13} and perform frequency selective filtering operations on graphs, similar to those in traditional signal processing.

Another very interesting smoothing filter is the guided filter (GF), recently proposed in
\cite{HST13,HS15}, and included into the MATLAB image processing toolbox. Some ideas behind GF
are developed in \cite{LLV08}. According to our limited experience, GF is faster than BF.
The authors of \cite{HST13} advocate that GF is gradient preserving and avoids gradient reversal
artifacts in contrast to BF, which is not gradient preserving.

The smoothing explicit filters similar to BF and GF can be interpreted as matrix power iterations,
which are, in general case, nonlinear, or equivalently, as explicit integration in time
of the corresponding nonlinear anisotropic diffusion equation \cite{PM87,PM90}.
The suitable graph Laplacian matrices
are determined by means of the graph-based interpretation of these power iterations.
Our main contribution is accelerating the smoothing filters by means of a special
variant of the conjugate gradient (CG) method, applied to the corresponding graph Laplacian matrices.
To avoid oversmoothing, only few iterations of the CG acceleration can be performed. 
We note that there exist several nonlinear variants of the CG algorithm, see, e.g., \cite{NW06}.
However, the developed theory is not directly applicable in our case because it is not clear how to
interpret the vector $L(x)x$ as a gradient of a scalar function of the signal $x$, where
$L(x)$ is a graph Laplacian matrix depending on a signal $x$.

\section{Bilateral filter (BF)}
\label{sec:BF}

We consider discrete signals defined on an undirected graph $\mathcal{G}=(\mathcal{V},\mathcal{E})$,
where the vertices $\mathcal{V}=\{1,2,\ldots,N\}$ denote, e.g., time instances
of a discrete-time signal or pixels of an image. The set of edges $\mathcal{E}=\{(i,j)\}$
contains only those pairs of vertices $i$ and $j$ that are neighbors in some predefined sense.
We suppose in addition that a spatial position $p_i$ is assigned to
each vertex $i\in\mathcal{V}$ so that a distance $\|p_i-p_j\|$ is determined between
vertices $i$ and $j$.

Let $x[j]$, $j\in\mathcal{V}$, be a discrete function, which is an input signal to the bilateral filter.
The output signal $y[i]$ is the weighted average of the signal values in $x[j]$:
\begin{equation}\label{eq1}
 y[i] = \sum_j\frac{w_{ij}}{\sum_jw_{ij}}x[j].
\end{equation}
The weights $w_{ij}$ are defined for $(i,j)\in\mathcal{E}$ in terms of a guidance signal $g[i]$:
\begin{equation}\label{eq2}
 w_{ij}=\exp\left(-\frac{\|p_i-p_j\|^2}{2\sigma_d^2}\right)
 \exp\left(-\frac{(g[i]-g[j])^2}{2\sigma_r^2}\right),
\end{equation}
where $\sigma_d$ and $\sigma_r$ are the filter parameters \cite{TM98}.
The guidance signal $g[i]$ is chosen depending on the purpose of filtering.
When $g$ coincides with the input $x$, the bilateral filter is nonlinear
and called self-guided.

The weights $w_{ij}$ determine the adjacency matrix $W$ of the graph $G$.
The matrix $W$ is symmetric, has nonnegative elements and diagonal elements equal to 1.
Let $D$ be the diagonal matrix with the nonnegative diagonal entries $d_i=\sum_{j}w_{ij}$.
Thus, the BF operation (\ref{eq1}) is the vector transform defined by the aid of the matrices
$W(g)$ and $D(g)$ as $y = D^{-1}Wx=x-D^{-1}Lx$, where $L=D-W$ is called the Laplacian matrix
of the weighted graph~$G$ with the BF weights. The eigenvalues of the matrix $D^{-1}W$ are real.
The eigenvalues corresponding to the highest oscillations lie near the origin.

The BF transform $y=D^{-1}Wx$ can be applied iteratively,
(i) by changing the weights $w_{ij}$ at each iteration using the result of the previous
iteration as a guidance signal $g$,
or (ii) by using the fixed weights, calculated from the initial signal as a guidance signal,
for all iterations.
The former alternative results in a nonlinear filter. The latter produces a linear filter,
which may be faster, since the BF weights are computed only once in the very beginning.

An iterative application of the BF matrix transform is the power iteration
with the amplification matrix $D^{-1}W$.
Slow convergence of the power iteration can be boosted
by the aid of suitable Krylov subspace iterative methods \cite{G97,V03}.

\section{Guided filter (GF)}
\label{sec:GF}

\vspace{1ex}
\begin{tabular}{l}
\hline\\[-2ex]
\textbf{Algorithm 1}{ Guided Filter (GF)}\\\hline\\[-2ex]
\textbf{Input:} $x$, $g$, $\rho$, $\epsilon$\\
\textbf{Output:} $y$\\
\quad $mean_g=f_{mean}(g,\rho)$\\
\quad $mean_x=f_{mean}(x,\rho)$\\
\quad $corr_g=f_{mean}(g.*g,\rho)$\\
\quad $corr_{gx}=f_{mean}(g.*x,\rho)$\\
\quad $var_g=corr_g-mean_g.*mean_g$\\
\quad $cov_{gx}=corr_{gx}-mean_{g}.*mean_x$\\
\quad $a = cov_{gx}./(var_g+\epsilon)$\\
\quad $b = mean_x-a.*mean_g$\\
\quad $mean_a=f_{mean}(a,\rho)$\\
\quad $mean_b=f_{mean}(b,\rho)$\\
\quad $y=mean_a.*g+mean_b$\\
\end{tabular}
\vspace{1ex}

Algorithm 1 is a pseudo-code of GF proposed in \cite{HST13}, where
$x$ and $y$ are, respectively, the input and output signals on the graph $G$, described
in section~\ref{sec:BF}. GF is built by means of a guidance signal $g$,
which equals $x$ in the self-guided case. The function $f_{mean}(\cdot,\rho)$
denotes a mean filter of a spatial radius $\rho$.
The constant $\epsilon$ determines the smoothness degree of the filter---the
larger $\epsilon$ the larger smoothing effect.
The dot preceded operations $.*$ and $./$ denote the componentwise multiplication and division.
A typical arithmetical complexity of the GF algorithm is $O(N)$,
where $N$ is the number of elements in $x$, see \cite{HST13}.

The guided filter operation of Algorithm 1 is the matrix transform $y = W(g)x$,
where the implicitly constructed transform matrix $W(g)$ has the following entries,
see \cite{HST13}: 
\begin{equation}\label{eq6}
W_{ij}(g)=\frac{1}{|\omega|^2}\sum_{k\colon (i,j)\in\omega_k}
\left(1+\frac{(g_i-\mu_k)(g_j-\mu_k)}{\sigma_k^2+\epsilon}\right).
\end{equation}
The mean filter $f_{mean}(\cdot,\rho)$ is applied in the neighborhoods $\omega_k$
of a spatial radius $\rho$ around all vertices $k\in\mathcal{V}$.
The number of pixels in $\omega_k$ is denoted by $|\omega|$, the same for all $k$.
The values $\mu_k$ and $\sigma_k^2$ are the mean and variance of $g$ over $\omega_k$.
The matrix $W$ is symmetric and satisfies the property $\sum_{j}W_{ij}=1$.

The standard construction of the graph Laplacian matrix gives $L=I-W$,
because $d_i=\sum_jw_{ij}=1$, i.e. the matrix $D$ is the identity.
The eigenvalues of $L(g)$ are real nonnegative with the low frequencies accumulated
near 0 and high frequencies near 1.
Application of a single transform $y=Wx$ attenuates the high frequency modes of $x$
while approximately preserving the low frequency modes, cf. \cite{GNO13,TKMV14}.
Similar to the BF filter, the guided filter can be applied iteratively. When the
guidance signal $g$ is fixed, the iterated GF filter is linear. When $g$ varies, for example,
$g=x$ for the self-guided case, the iterated GF filter is nonlinear.

\section{Conjugate gradient acceleration}
\label{sec:bpcg}

Since the graph  Laplacian matrix $L$ is symmetric and nonnegative definite,
the iterative application of the transform $y=D^{-1}Wx$ can be accelerated
by adopting by the CG technology.
We use two variants of CG:
1) with the fixed guidance equal to the input signal or to the clean signal,
2) with the varying guidance equal to the current value of $x$.

\vspace{1ex}
\begin{tabular}{l}
\hline\\[-2ex]
\textbf{Algorithm 2}{ Truncated PCG($k_{\max}$)}\\\hline\\[-2ex]
\textbf{Input:} $x_0$, $g$, $k_{\max}$ \textbf{ Output:} $x$\\
$x=x_0$; $r=W(g)x-D(g)x$\\
\textbf{for }{$k=1,\ldots,k_{\max}-1$}\textbf{ do}\\
\quad $s=D^{-1}(g)r$; $\gamma=s^Tr$\\
\quad\textbf{if }$k=1$\textbf{ then} $p=s$ \textbf{else}
$\beta=\gamma/\gamma_{old}$; $p=s+\beta p$\\
\quad\textbf{endif}\\
\quad $q=D(g)p-W(g)p$; $\alpha=\gamma/(p^Tq)$\\
\quad $x=x+\alpha p$; $r=r-\alpha q$; $\gamma_{old}=\gamma$\\
\textbf{endfor}\\
\end{tabular}
\vspace{1ex}

Algorithm 2 is the standard preconditioned conjugate gradient algorithm formally applied
to the system of linear equations $Lx=0$ and truncated after $k_{\max}$ evaluations of
the matrix-vector operation $Lx$. The initial vector $x_0$ is a noisy input signal. 
This variant of the CG algorithm has first been suggested in \cite{TKMV14}.

Algorithm 3 is a special nonlinear preconditioned CG with $l_{\max}$ restarts,
formally applied to $L(x)x=0$ and truncated after $k_{\max}$ iterations
between restarts. Restarts are necessary because of nonlinearity of the
self-guided filtering.

\vspace{1ex}
\begin{tabular}{l}
\hline\\[-2ex]
\textbf{Algorithm 3}{ Truncated PCG($k_{\max}$) with $l_{\max}$ restarts}\\\hline\\[-2ex]
\textbf{Input:} $x_0$, $k_{\max}$, $l_{max}$ \textbf{ Output:} $x$\\
$x=x_0$\\
\textbf{for }{$l=1,\ldots,l_{\max}$}\textbf{ do}\\
\quad $r=W(x)x-D(x)x$\\
\quad\textbf{for }{$k=1,\ldots,k_{\max}-1$}\textbf{ do}\\
\qquad $s=D^{-1}(x)r$; $\gamma=s^Tr$\\
\qquad\textbf{if }$k=1$\textbf{ then} $p=s$
\textbf{else} $\beta=\gamma/\gamma_{old}$; $p=s+\beta p$\\
\qquad\textbf{endif}\\
\qquad $q=D(x)p-W(x)p$; $\alpha=\gamma/(p^Tq)$\\
\qquad $x=x+\alpha p$; $r=r-\alpha q$; $\gamma_{old}=\gamma$\\
\quad\textbf{endfor}\\
\textbf{endfor}\\
\end{tabular}
\vspace{1ex}

\section{Numerical experiments}
\label{sec:exper}

\begin{figure}[!t]
\centering
\includegraphics[width=\columnwidth,height=25ex]{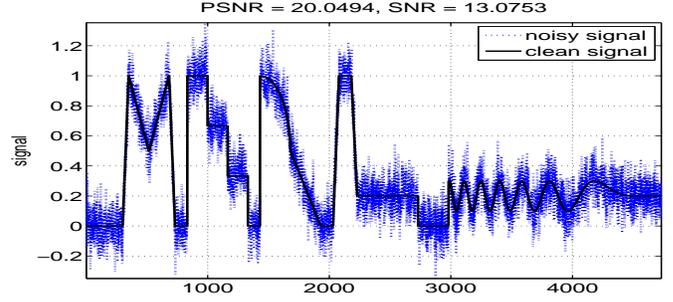}
\caption{Clean and noisy signals.}
\label{pic0}
\end{figure}

As a proof of concept, our MATLAB tests use the clean 1-dimensional signal $x_c$
of length $N=4730$ shown in Figure~\ref{pic0}.
We choose this rather difficult, although 1-dimensional, example to better visually illustrate
both the denoising and edge-preserving features of the filters.
The noisy signal, also displayed in Figure~\ref{pic0},
is the same for all tests and given by the formula $x_0=x_c+\eta$, where a Gaussian
white noise $\eta$ has zero mean and variance $\sigma^2=0.01$. The bilateral filter
is used with $\sigma_d=0.5$ and $\sigma_r=0.1$.
The neighborhood width in BF equals 5 so that the band of $W$ consists of 5 diagonals.
The guided filter is used with $\epsilon=0.001$ and the neighborhood width 3.
The matrix $W$ of GF also has 5 diagonals.

The CG accelerated BF/GF is called CG-BF/CG-GF.
Typical numerical results of the average performance are displayed. 

The signal error after denoising is $\widehat{x}-x_0$, where $\widehat{x}$ stands for the output
denoised signal. We calculate the peak signal-to-noise ratio (PSNR) and signal-to-noise ration (SNR).
The~parameters are manually optimized to reach the best possible
match of the signal errors in the compared filters,
resulting in indistinguishable error curves in our figures.

\begin{figure}[!t]
\centering
\includegraphics[width=\columnwidth]{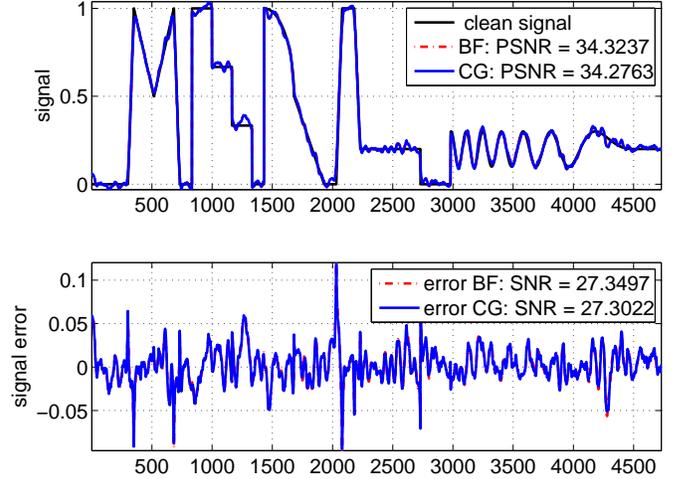}
\caption{500 BF iterations versus 20 CG-BF iterations with the guidance $x_c$.}
\label{pic1}
\end{figure}

\begin{figure}[!t]
\centering
\includegraphics[width=\columnwidth]{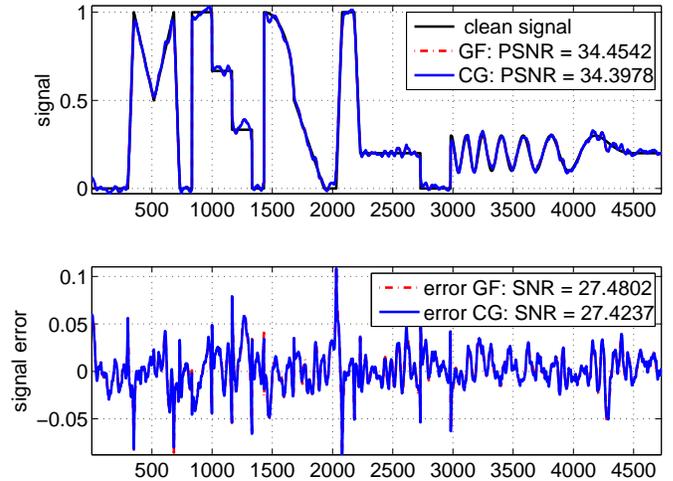}
\caption{90 GF iterations versus 13 CG-GF iterations with the guidance $x_c$.}
\label{pic2}
\end{figure}

The results in Figures~\ref{pic1} and \ref{pic2} are obtained by the iterated
BF and GF filters with the fixed guidance $g=x_c$ and by CG-BF and CG-GF implemented
in Algorithm~2 with the same fixed guidance $g=x_c$. These tests are performed only
for comparison reasons because the clean signal guidance $x_c$ seems to be ideal for
the best possible denoising results.

\begin{figure}[!t]
\centering
\includegraphics[width=41em]{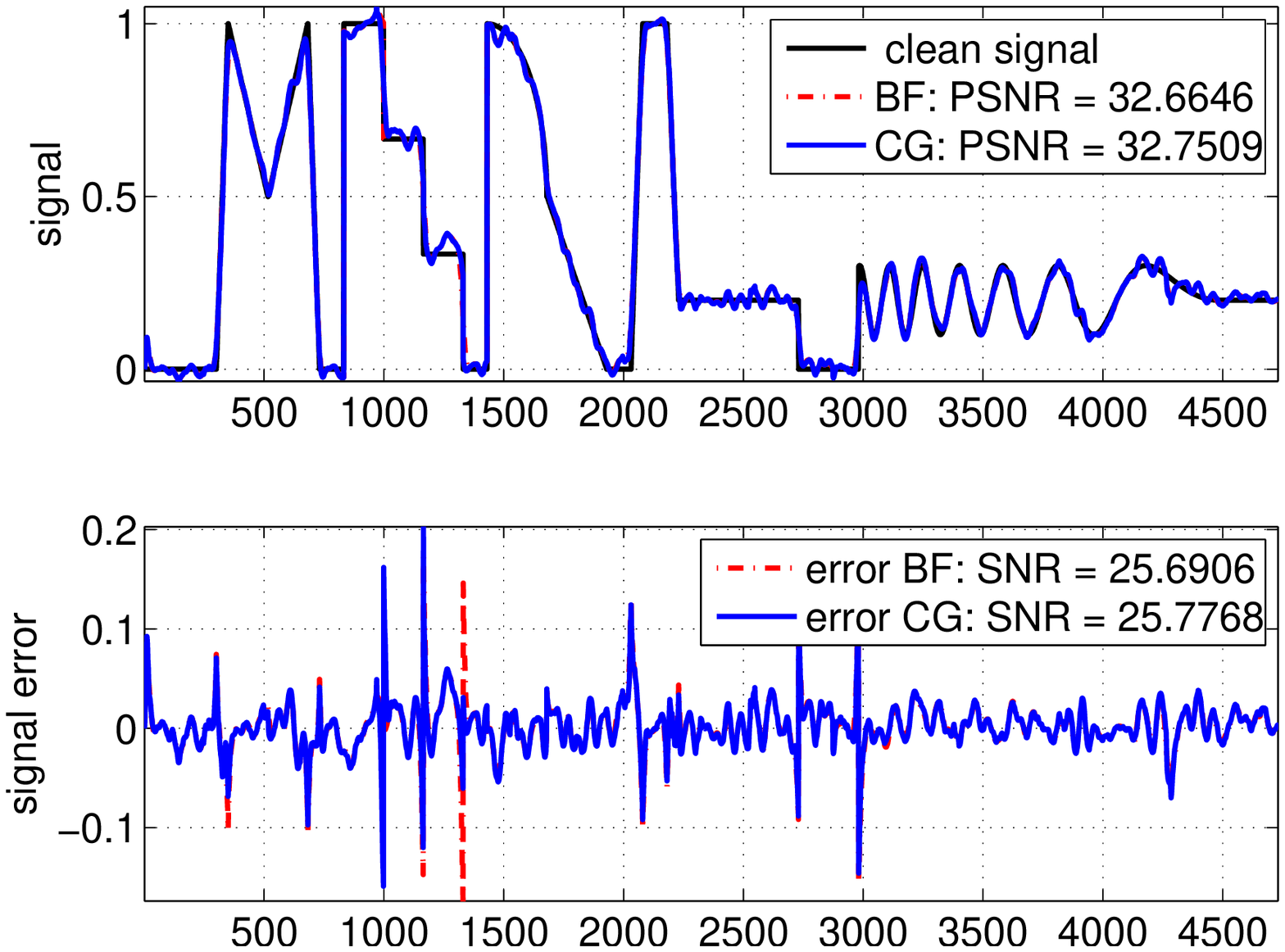}
\caption{600 iterations of the self-guided BF versus $3\times11$ iterations of CG-BF.}
\label{pic3}
\end{figure}

\begin{figure}[!t]
\centering
\includegraphics[width=41em]{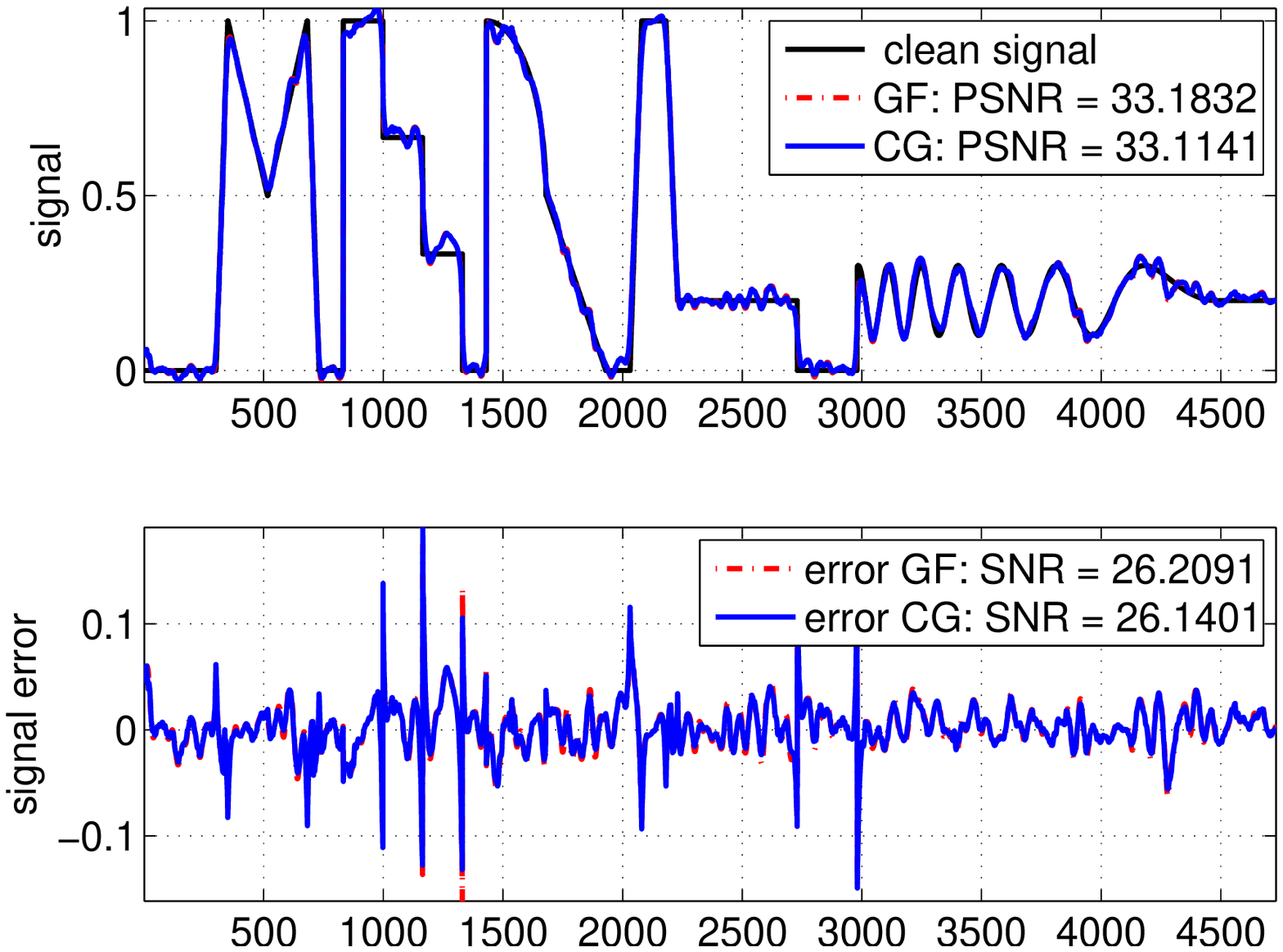}
\caption{75 iterations of the self-guided GF versus $5\times5$ iterations of CG-GF.}
\label{pic4}
\end{figure}

We say that Algorithm 3 uses $l_{\max}\times k_{\max}$ iterations, if it executes $l_{\max}$ restarts
with the $k_{\max}$ evaluations $L(x)x$ between restarts. The best denoising performance for our
test problem is achieved with the following iteration combinations of the self-guided CG-BF:
$31\times3$, $17\times4$, $12\times5$, $9\times6$, $7\times7$, $6\times8$, $5\times9$, $4\times10$, $3\times11$, $2\times19$. The best combinations for the self-guided CG-GF are $11\times3$, $7\times4$,
$5\times5$, $4\times6$, $3\times7$. Figures~\ref{pic3} and \ref{pic4} show the results
after $3\times11$ iterations of CG-BF and $5\times5$ iterations of CG-GF.

The numerical tests demonstrate about $20$-times reduction of iterations
for the self-guided bilateral filter and $3$-times reduction of iterations
for the guided filter with self-guidance after the conjugate gradient acceleration.
It is also interesting to observe that both filters with the properly chosen parameters
and iteration numbers produce almost identical output signals.

\section{Conclusion}

Iterative application of BF and GF, including
their nonlinear self-guided variants, can be drastically accelerated by using CG
technology.
Our future work concerns developing automated procedures for choosing the optimal number
of CG iterations and investigating CG acceleration for 2D signals.

\onecolumn

\twocolumn



\IEEEtriggeratref{8}


%

\pagebreak

\end{document}